\documentclass[letterpaper, 10 pt, conference]{ieeeconf}  
\IEEEoverridecommandlockouts                              

\usepackage{graphicx} 
\usepackage{times} 
\usepackage{cite}
\usepackage{amsmath} 
\usepackage{amssymb}  
\usepackage{amsfonts}       
\usepackage{dsfont}
\usepackage{braket}
\usepackage{booktabs}
\usepackage{color}
\usepackage{colortbl}
\usepackage{url}
\usepackage{mathtools}
\usepackage{multirow}
\usepackage{footmisc}
\usepackage{mwe}
\usepackage[linesnumbered,ruled,vlined]{algorithm2e}
\SetKwInput{KwInput}{Input}                
\SetKwInput{KwOutput}{Output}

\SetCommentSty{mycommfont}

\usepackage[caption=false, font=footnotesize]{subfig}

\usepackage{lipsum}
\usepackage{widetext}
\newcommand{\obs}{\boldsymbol{x}}
\newcommand{\inp}{\theta}
\newcommand{\outp}{y} 
\newcommand{\argmax}{\mathop{\rm argmax}\limits}

\newcommand{\state}{\mathbf{x}}
\newcommand{\action}{\mathbf{u}}
\newcommand{\vel}{\dot{\obs}}
\newcommand{\inertia}{{\Lambda}}
\newcommand{\dampcoef}{{D}}
\newcommand{\stiffness}{{K}}
\newcommand{\force}{\boldsymbol{F}}

\newcommand{\best}[2]{\underline{\textbf{#1}}$\pm$\scriptsize{#2}}
\newcommand{\second}[2]{\underline{#1} $\pm$ \scriptsize{#2}}
\newcommand{\third}[2]{#1$\pm$\scriptsize{#2}}

\title{\LARGE \textbf{
  Learning Compliant Stiffness by
  Impedance Control-Aware Task Segmentation and Multi-objective Bayesian Optimization with Priors
}}

\author{Masashi Okada$^{\dag,\star}$, Mayumi Komatsu$^{\ddag}$, Ryo Okumura$^{\dag}$ and Tadahiro Taniguchi$^{\dag,*}$
\thanks{$^{\dag}$ Masashi Okada, Ryo Okumura and Tadahiro Taniguchi are with Digital \& AI Technology Center, Technology Division, Panasonic Holdings Corporation, Japan.
}%
\thanks{$^{\ddag}$ Mayumi Komatsu is with Robotics Promotion Office, Manufacturing Innovation Division, Panasonic Holdings Corporation, Japan.
}%
\thanks{$^{*}$ Tadahiro Taniguchi is also with Ritsumeikan University, College of Information Science and Engineering, Japan.
}%
\thanks{$^{\star}$ \texttt{okada.masashi001@jp.panasonic.com}
}
}

\begin{document}

\maketitle
\thispagestyle{empty}
\pagestyle{empty}

\begin{abstract}
Rather than traditional position control, impedance control is preferred to ensure the safe operation of industrial robots programmed from demonstrations.
However, variable stiffness learning studies have focused on task performance rather than safety (or compliance).
Thus, this paper proposes a novel stiffness learning method to satisfy both task performance and compliance requirements.
The proposed method optimizes the \textit{task and compliance objectives (T/C objectives)} simultaneously via multi-objective Bayesian optimization.
We define the stiffness search space by segmenting a demonstration into task phases,
each with constant responsible stiffness.
The segmentation is performed by identifying \textit{impedance control-aware switching linear dynamics} (IC-SLD) from the demonstration.
We also utilize the stiffness obtained by proposed IC-SLD as priors for efficient optimization.
Experiments on simulated tasks and a real robot demonstrate that IC-SLD-based segmentation and the use of priors improve the optimization efficiency compared to existing baseline methods.
\end{abstract}

\section{Introduction} \label{sec:intro}
Learning from demonstration (LfD), especially robot control based on the playback of demonstrated trajectories, is an accepted technology in industry
thanks to its intuitiveness and the ease of implementation of its simple feedback controller in low-cost embedded systems.
This study focuses on extending this control scheme to contact-rich and/or human collaborative tasks without requiring significant updates to existing systems.

To achieve safe and satisfactory task performance, it is essential to introduce variable impedance control with appropriately designed stiffness~\cite{7110619,7560657}.
Generally, cartesian position control is used for the demonstration playback;
however, this can potentially cause damage to the robot or its surroundings
due to unforeseen contact.
In contrast, impedance control makes the robot's behavior compliant with respect to external forces, thereby ensuring safety.
Stiffness affects the safety and reproducibility of the demonstration,
and there is a tradeoff between the two objectives~\cite{pollayil2022choosing}.
Thus, we can formulate the stiffness determination problem as the optimization of two objective functions, i.e., the \textit{task objective} (e.g., tracking error and sparse reward indicating task success) and \textit{compliance objective} (e.g., a penalty for high stiffness).
\begin{figure}[t]
  \centering
  \includegraphics[width=0.45\textwidth]{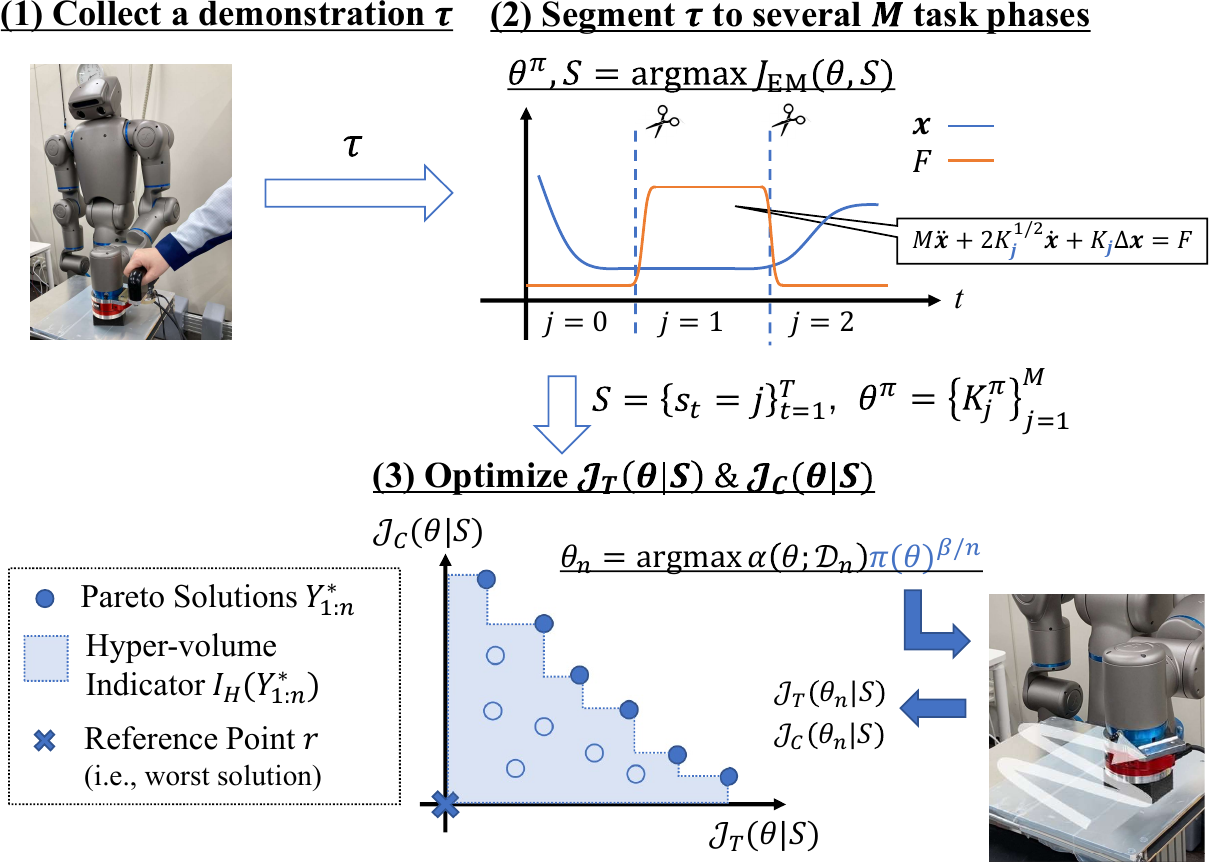}
  \caption{
    Overview of the proposed to learn stiffness parameters.
    (1) A human demonstration $\tau$ is collected by direct teaching or teleoperation.
    (2) The demonstration $\tau$ is fitted to the \textit{impedance control-aware switching linear dynamics model} (IC-SLD) to segment $\tau$ into several task phases,
    resulting in the phase information $S=\{s_{t}=j\}$.
    This step also estimates the stiffness $\theta^{\pi} = \{\stiffness^{\pi}_{j}\}$ in the segmented phases.
    (3) Multi-objective optimization of \textit{task and compliance objectives} (denoted as $\mathcal{J}_{T}(\theta|S)$ and $\mathcal{J}_{C}(\theta|S)$, respectively) is performed by 
    testing stiffness in real environments.
    The Bayesian optimization method suggests the control parameters $\theta_{n}$ to test,
    where the estimated $\theta^{\pi}$ are employed as the prior for $\pi$-BO~\cite{hvarfner2022pi}.
  } \label{fig:fig1}
\end{figure}

In previous studies on impedance control,
simultaneous optimization of the task and compliance objectives (\textit{T/C objectives}) have yet to be
investigated extensively, thereby raising safety concerns.
For example, previous studies~\cite{johannsmeier2019framework,wu2022prim} optimized only the task objective using a Bayesian optimization process to identify the stiffness that leads to task success.
In contrast, other studies~\cite{pollayil2022choosing,lukic2022online} optimized the compliance objective but neglected the task objective.

Simultaneous optimization of multiple objectives frequently requires reward engineering~\cite{dewey2014reinforcement},
which is particularly problematic when the optimization problem involves \textit{online} testing with real robots.
A naive method to optimize T/C objectives simultaneously is to derive the scalarized objective by a weighted linear combination~\cite{buchli2011learning};
however, this method requires iterative tuning of the weights and optimization to find the desired behavior.
In addition, the task objective should include a sparse reward because the user's primary goal is task success. 
This type of task objective is challenging to predict using a physical model and optimize with gradient methods; thus, testing the robots on actual equipment is needed.
Even if the model can predict the task objective effectively, online testing procedures are necessary because a gap exists between the model and the real-world environment.

Multi-objective optimization and Bayesian optimization are promising solutions to the above problems.
Multi-objective optimization has several benefits, e.g., 
it provides multiple optimal solutions (\textit{Pareto solutions}) that represent the best tradeoff in a single optimization process without reward engineering.
The following two concepts can also contribute to online Bayesian optimization:
\textit{(1) appropriate segmentation of demonstrations} and \textit{(2) use of stiffness priors.}
Bayesian optimization performs poorly in high dimensions.
For concept (1), rather than finding varying stiffness at each time step,
the input dimension is reduced by segmenting a task into several phases and assigning constant responsible stiffness values to each phase.
In previous studies, the segmentation was performed manually~\cite{johannsmeier2019framework,9981728,wu2022prim} or by using the Gaussian Mixture Model (GMM) ~\cite{ENAYATI20209834,le2021learning};
however, these segmentation methods are not designed to optimize T/C objectives. 
For concept (2), the stiffness estimated by model-based methods can be a good candidate solution.
Unfortunately, standard Bayesian optimization cannot incorporate prior knowledge~\cite{souza2020prior}.

This paper substantiates and validates these two concepts.
Figure~\ref{fig:fig1} summarizes the top-level concept.
Our primary contributions are summarized as follows.
\begin{itemize}
  \item We propose to learn stiffness via multi-objective optimization of the T/C objectives.
  \item We introduce an \textit{impedance control-aware switching linear dynamics model} (IC-SLD) that effectively segments a demonstration and identifies stiffness.
  \item We employ the state-of-the-art Bayesian optimization method $\pi$-BO~\cite{hvarfner2022pi} to utilize the estimated stiffness as the priors of a promising solution candidate.
\end{itemize}

The remainder of this paper is organized as follows.
In Sec.~\ref{sec:related_work}, we summarize related work.
In Sec.~\ref{sec:preliminary}, we briefly review preliminaries, i.e., impedance control, Bayesian and multi-objective optimization, and switching linear dynamics (SLD).
Sec.~\ref{sec:method} presents the proposed stiffness learning method,
and the effectiveness of the proposed method is demonstrated in Sec.~\ref{sec:experiments}. 
Finally, the paper is concluded in Sec.~\ref{sec:conclusion}.

\section{Related Work} \label{sec:related_work}
\subsection{Learning Stiffness with Interaction} \label{sec:choosing_stiffness}
Various techniques have been proposed in the literature on variable impedance control to learn stiffness~\cite{abu2020variable}.
In the following, we review the learning methods involving online robot testing.

Reinforcement learning (RL) is a representative method of determining stiffness through real-world interaction.
In a previous study, Buchli applied Policy Improvement with Path Integral (PI$^{2}$) to tune the stiffness at each timestep~\cite{buchli2011learning}.
In addition, previous studies~\cite{martin2019variable,bogdanovic2020learning} trained neural policies using RL to infer stiffness and attractors from current observations.
Recent studies~
\cite{oikawa2021reinforcement,kozlovsky2022reinforcement} have attempted to learn neural policies to determine nondiagonal stiffness matrices.
However, RL remains sample inefficient and requires numerous interactions in real-world environments.

Another strategy to learn control parameters in an online manner is using black-box optimization.
In several previous studies, various black-box optimization methods have been applied to stiffness learning, e.g.,
Bayesian optimization~\cite{johannsmeier2019framework,wu2022prim},
Covariance Matrix Adaptation Evolutionary~\cite{8566177}, and
Particle Swarm Optimization~\cite{salehi2008impedance,fateh2011adaptive,azimi2015stable}.

\subsection{Segmentation of Demonstrations} \label{sec:segmentation}
In robotics, segmentation has long been a primary research subject due to its attractive applications, e.g., reusable skill discovery and hierarchical reinforcement learning.
Accordingly, various methods have been proposed, e.g.,
Fourier basis-based segmentation~\cite{konidaris2012robot},
clustering by GMM~\cite{calinon2016tutorial,krishnan2017transition},
movement matching from a pre-existing skill set of Dynamic Movement Primitives~\cite{meier2011movement}, simultaneous segmentation and learning of Probabilistic Movement Primitives~\cite{lioutikov2017learning},
and SLD-based system identification~\cite{murali2016tsc,abdulsamad2020hierarchical}.
In addition, deep RL has recently focused on acquiring neural skills by training multiple policies conditioned on segment identifiers
\cite{shiarlis2018taco,shankar2019discovering,kipf2019compile,shankar2020learning,tanneberg2021skid,villecroze2022bayesian}.

\subsection{Multi-objective Optimization in Robotics} \label{sec:multiobj_in_robot}
Designing robot mechanisms or controllers while satisfying conflicting objectives is a popular topic in robotics,
e.g., the speed vs. head stability of the gait of a snake robot~\cite{6630691},
the initial vs. running costs of industrial robot arms~\cite{kouritem2022multi},
and the desired joint trajectory vs. regenerate the energy of a prosthetic leg~\cite{7799085}.
A recent study \cite{9926713} proposed a method to optimize both performance and safety metrics using $\pi$-BO, where the safety metric was the distance between the robot and fragile items, and users defined the priors.

\section{Preliminaries} \label{sec:preliminary}
\subsection{Cartesian Impedance Control} \label{sec:imp_ctrl}
The purpose of impedance control is to impose the robot's dynamics to follow the closed-impedance model:
\begin{align}
  & \inertia \Delta \ddot{\obs} + \dampcoef \Delta \vel + \stiffness \Delta \obs = \force \label{eqn:impedance_control}, \\
  & \Delta \obs = \obs_{d} - \obs,
\end{align}
where $\obs \in \mathbb{R}^{6}$ is the end-effector position and orientation in task space,
$\obs_{d} \in \mathbb{R}^{6}$ is the attractor, 
$\force \in \mathbb{R}^{6}$ is the external force/torque acting on the end-effector,
and the matrices
$\inertia \in \mathbb{R}^{6\times 6}$,
$\dampcoef \in \mathbb{R}^{6\times 6}$,
and $\stiffness \in \mathbb{R}^{6\times 6}$ are 
the desired Cartesian inertia, damping, and stiffness, respectively.
Here, the spring behavior realized in terms of stiffness $\stiffness$ allows the robot to follow
the desired trajectory of the attractor $\obs_{d}$ 
while making the robot flexible to unexpected external forces $\force$.
To make the system compliant, stiffness $\stiffness$ should be as low as possible.
In addition, damping $\dampcoef$ is frequently set in proportion to the square root of the stiffness~\cite{robosuite2020,1242165,9829927,9583671};
\begin{align}
  \dampcoef = 2 \stiffness^{\frac{1}{2}} \label{eqn:critically_dumped}.
\end{align}
Thus, in the current study, $\dampcoef$ is not subject to optimization.

\subsection{Bayesian and Multi-objective Optimization} \label{sec:bayesian_opt}
\subsubsection{Basis}
We consider the following optimization problem of function $f$ across a set of feasible input $\Theta$;
\begin{align}
  \mathbf{\inp}^{*} = \argmax_{\theta \in \Theta} f(\inp),
\end{align}
where $\mathbf{\inp}^{*}$ denotes the optimal solution.
%
Bayesian optimization is an iterative process involving a candidate solution suggested by an acquisition function $\alpha(\inp; \mathcal{D}_{n})$:
\begin{align}
  \inp_{n + 1} = \argmax_{\theta \in \Theta} \alpha(\inp; \mathcal{D}_{n}),
\end{align}
where $n$ is the number of iterations, and $\mathcal{D}_{n} = \{(\inp_{i}, \outp_{i}=f(\inp_{i}))\}_{i=1}^{n}$ is the dataset of assessed candidates.
Several acquisition functions have been proposed previously,
and the most common is Expected Improvement (EI)~\cite{jones1998efficient}:
\begin{align}
  \alpha(\inp; \mathcal{D}_{n}) = \mathbb{E}_{p(\outp|\inp)}\left[\max(y - y^{*}_{1:n}, 0)\right]
\end{align}
where $\outp^{*}_{1:n}$ is the best objective value observed by iteration $n$.
However, the oracle of $p(\outp|\inp)$ is unknown; thus, it is necessary to evaluate $\alpha$ via surrogate modeling by Gaussian processes~\cite{jones1998efficient} and a tree-structured Parzen estimator~\cite{NIPS2011_86e8f7ab}.

\subsubsection{Multi-objective Optimization} \label{sec:multiobj_opt}
For brevity, we consider the optimization problem of two functions $f_{1}$ and $f_{2}$ without loss of generality:
\begin{align}
  \mathbf{\inp}^{*} = \argmax_{\theta \in \Theta} f_{1}(\inp), f_{2}(\inp).
\end{align}
Generally, these objectives conflict with each other.

Here, let $Y_{1:n}^{*}$ be the set of Pareto solutions observed by iteration $n$, representing the best tradeoff between the target objectives.
The current state-of-the-art multi-objective Bayesian optimization technique~\cite{ozaki2020multiobjective} defines EI such that \textit{the hypervolume indicator} $I_{H}(Y_{1:n}^{*})$ (illustrated in Fig.~\ref{fig:fig1}) is improved:
\begin{align}
  \alpha(\inp; \mathcal{D}_{n}) = \mathbb{E}_{p(y|\theta)}\left[\max(I_{H}(Y_{1:n}^{*} \cup \{y\}) - I_{H}(Y_{1:n}^{*}),0)\right], \label{eqn:alpha_multi_obj}
\end{align}
where, in the case of two objectives,  $I_{H}(Y_{1:n}^{*})$ is the area composed of $Y_{1:n}^{*}$ and a \textit{reference point} $r$:
\begin{align}
  I_{H}(Y) \coloneqq \lambda\left(\bigcup_{\outp \in Y}  [y_{1}, r_{1}] \times [y_{2}, r_{2}]\right). \label{eqn:hypervolume}
\end{align}
Here, $\lambda(S)$ is the area of a set $S$, and $[y_{1}, r_{1}] \times [y_{2}, r_{2}]$ represents a rectangle comprised of the two edges.

\subsubsection{Bayesian Optimization with Priors}
Conventional Bayesian optimization methods cannot incorporate prior knowledge
other than narrowing the search space. However, this hard prior can result in suboptimal performance because of missing important regions.
Recently proposed $\pi$-BO incorporates the prior in the form of a probability distribution $\pi(\theta)$ into an optimum, and it utilizes the following decaying prior-weighted acquisition function:
\begin{align}
  \alpha_{\pi}(\inp; \mathcal{D}_{n}) \coloneqq \alpha(\inp; \mathcal{D}_{n}) \pi(\inp)^{\beta / n}, \label{eqn:pibo}
\end{align}
where $\beta \in \mathbb{R}^{+}$ is a hyperparameter reflecting the confidence in $\pi(\theta)$.
Initially, the acquisition function gives significant weight to the prior;
however, with increasing $n$, the exponent of the prior decreases gradually toward zero, thereby making $\alpha_{\pi}$ similar to $\alpha$.

\subsection{Switching Linear Dynamics} \label{sec:switching_linear_dyn}
Switching linear dynamics (SLD) models a system as a collection of linear dynamics,
where each model represents an operating mode~\cite{murphy1998switching}.
Here, let $\tau = (\state_{1},\action_{1}, \cdots, \action_{T-1}, \state_{T})$ be an observed trajectory
comprising of states $\state$ and control inputs $\action$, which is assumed to be generated by the following stochastic dynamics:
\begin{align}
  & p(\state_{1:T}|\action_{1:T}, s_{1:T}) = \prod^{T-1}_{t=1} p(\state_{t+1}|\state_{t},\action_{t}, s_{t}=j),
\end{align}
where $p(\state_{t+1}|\cdot)$ is the Gaussian linear model:
\begin{align}
  p(\state_{t+1}|\state_{t},\action_{t}, s_{t}=j) &\coloneqq \mathcal{N}(\state_{t+1}; A_{j}\state_{t} + B_{j}\action_{t}, \Sigma_{j}), \label{eqn:gaussian_linear}
\end{align}
Here, $s_{t} \in \{1,2,\cdots,M\}$ is the \textit{discrete hidden switch variable} (or segment identifier),
$M$ is the number of linear models,
and $A_{j}$, $B_{j}$, and $\Sigma_{j}$ are the dynamics parameters depending on the switch state $s_{t} = j$.
In this task, the goal is to identify the dynamics parameters $\theta \coloneqq \{A_{j}, B_{j}, \Sigma_{j}\}_{j=1}^{M}$ and infer the hidden states $S\coloneqq \{s_{t}\}_{t=1}^{T}$,
which is performed by maximizing the following objective:
\begin{align}
  \mathcal{J}_{\mathrm{EM}}(\theta, S) &\coloneqq \sum_{t=1}^{T-1} \sum_{j=1}^{M} W_{t}^{j} \cdot \mathcal{J}^{j}_{t}, \label{eqn:obj_em} \\
  W_{t}^{j} &\coloneqq p(s_{t} = j  | \state_{1:T}, \action_{1:T}), \\
  \mathcal{J}^{j}_{t} &\coloneqq \log p(\state_{t+1}|\state_{t},\action_{t}, s_{t}=j).
  \label{eqn:em_obj_jt}
\end{align}
We can solve this optimization problem numerically by the expectation-maximization algorithm (EM) algorithm~\cite{murphy1998switching},
which conducts the following \textit{E-step} and \textit{M-step} iteratively until convergence.
Here, the \textit{E-step} calculates $W_{t}^{j}$ with fixed $\theta$,
and \textit{M-step} updates $\theta$ by maximizing Eq.~(\ref{eqn:obj_em}) with fixed $W_{t}^{j}$.
%

%


\section{Method} \label{sec:method}
\subsection{Problem Statement}
We consider dividing a demonstrated trajectory $\tau$ into $M$ segments (or task phases)
and assigning a constant stiffness $\stiffness_{j}$ to each segment $j \in \{1,2,\cdots,M\}$.
Here, let $S=\{s_{t}=j\}_{t=1}^{T}$ and $\theta = \{\stiffness_{j}\}_{j=1}^{M}$ be
the set of segment identifiers ($s_{t} \in \{1,2,\cdots,M\}$) and the parameter set, respectively.
The target multi-objective optimization problem is formulated as follows:
\begin{align}
  \argmax_{\theta \in \Theta} \mathcal{J}_{T}(\theta | S), \mathcal{J}_{C}(\theta | S), \label{eqn:mult_obj_opt}
\end{align}
where $\mathcal{J}_{T}(\theta | S)$ and $\mathcal{J}_{C}(\theta | S)$ are the task and compliance objective, respectively.
Here, $\mathcal{J}_{C}(\theta | S)$ is defined as follows:
\begin{align}
  \mathcal{J}_{C}(\theta | S) \coloneqq - \sum_{t=1}^{T} |\stiffness_{s_{t}}|.
\end{align}
This objective sums the penalties for high stiffness at each time step.
The task objective $\mathcal{J}_{T}$ assesses the task performance as follows:
\begin{align}
  \mathcal{J}_{T}(\theta | S) \coloneqq \sum_{t=1}^{T} R(\obs_{t}),
\end{align}
where $R$ is a task-specific reward function to evaluate each state $\obs_{t}$,
and the state transitions are dominated by $\stiffness_{s_{1:T}}$.
Note that both $\mathcal{J}_{T}$ and $\mathcal{J}_{C}$ are highly influenced by $S$;
however, $S$ is not the optimization target in Eq.~(\ref{eqn:mult_obj_opt}).
Thus, we must select $S$ carefully to realize effective optimization.

\subsection{Impedance Control-aware Switching Linear Dynamics}
The application of SLD is reasonable for the above setup of assigning constant parameters to the task phases (or \textit{switching stiffness control}). Here, we introduce \textit{impedance control-aware switching linear dynamics} (IC-SLD), which incorporates the impedance model priors from Eq.~(\ref{eqn:impedance_control}) to formulate the SLD identification problem.
With this impedance control-aware formulation, we expect to identify task phases suitable for the switching stiffness control performed during the subsequent optimization.

Here, the state $\state_{t}$ and action $\action_{t}$ are defined as
$\state_{t} \coloneqq (\dot{\obs_{t}}, \obs_{t}) \in \mathbb{R}^{12}$ and $\action_{t} \coloneqq (\Delta \obs_{t}, \force_{t}) \in \mathbb{R}^{12}$, respectively,
where $\Delta \obs_{t}$ denotes the residual from the attractor, which we regard as control inputs.
Note that we cannot extract $\Delta \obs_{t}$ from the human demonstration;
thus we assume that $\Delta \obs_{t} \coloneqq \obs_{t+1} - \obs_{t}$ during the segmentation step.
We also assume that sensors obtain the force observations during the demonstration.
According to these definitions and by discretizing Eq.~(\ref{eqn:impedance_control}) with the Euler method, 
we can specify the linear dynamics parameters $A^{j}$ and $B^{j}$ in Eq.~(\ref{eqn:gaussian_linear}) as follows:
\begin{align}
  A_{j} &=
  \begin{pmatrix}
    I - \inertia^{-1} \cdot 2\stiffness_{j}^{\frac{1}{2}} \Delta t & O \\
    I\cdot \Delta t & I
  \end{pmatrix}, \label{eqn:matrixA}\\ 
  B_{j} &=
  \begin{pmatrix} 
    \inertia^{-1} \stiffness_{j} \Delta t & \inertia^{-1} \Delta t \\
    O & O
  \end{pmatrix}, \label{eqn:matrixB} 
\end{align}
where $\Delta t$ is the sampling period. We also assume that $D=2\stiffness^{1/2}$, as in Eq.~(\ref{eqn:critically_dumped}).
By substituting Eqs.~(\ref{eqn:matrixA}) and (\ref{eqn:matrixB}) into Eqs.~(\ref{eqn:em_obj_jt}) and  (\ref{eqn:gaussian_linear}), we obtain the following:
\begin{align}
  \mathcal{J}^{j}_{t} \propto - \delta \state^{\mathsf{T}}_{t} \Sigma^{-1}_{j} \delta \state_{t} - \log |\Sigma_{j}|, \label{eqn:em_obj_jt2}
\end{align}
where $\delta \state_{t} = ( \delta \vel_{t}, \delta \obs_{t})$, and
\begin{align}
  & \delta \vel_{t} \coloneqq \vel_{t+1} - \vel_{t} - \inertia^{-1}(\stiffness_{j} \Delta \obs_{t} + 2{\stiffness_{j}^{\frac{1}{2}}} \vel_{t} - \force_{t}) \Delta t, \\
  & \delta \obs_{t} \coloneqq \obs_{t+1} - \obs_{t} - \vel_{t} \Delta t.
\end{align}
Generally, demonstrations do not include the observation of velocity $\vel_{t}$;
thus, the velocity is approximated as $\vel_{t} \simeq (\obs_{t} - \obs_{t-1})/\Delta t$.
In this case, $\delta \obs_{t} = \mathbf{0}$ and the related term in Eq.~(\ref{eqn:em_obj_jt2}) can be ignored.
In terms of $\sigma_{j}$, we found that relating the variance matrix to the stiffness matrix stabilizes the optimization:
\begin{align}
  \Sigma_{j}[1{:}6; 1{:}6] = \kappa \stiffness_{j},
\end{align}
where $\kappa \in \mathbb{R}^{+}$ is a hyperparameter described later in this section.
Consequently, we utilize the following objective for the EM algorithm:
\begin{align}
  \mathcal{J}^{j}_{t} \propto - \delta \vel_{t}^{\mathsf{T}} \stiffness^{-1}_{j} \delta \vel_{t} - \kappa \log|\stiffness_{j}|. \label{eqn:em_obj_jt3}
\end{align}
The first term of this objective is to find $\stiffness_{j}$ that minimizes the discrepancy from the motion equation of Eq.~(\ref{eqn:impedance_control}).
In addition, the term can be optimized by increasing the stiffness (or decreasing $\stiffness^{-1}_{j}$) if the discrepancy cannot be solved due to noisy observation $\force_{t}$, but this behavior is penalized by the second term weighted by the hyperparameter $\kappa$.
In a previous study~\cite{ENAYATI20209834}, we observed a similar concept of relating variance and stiffness.
However, with this formulation, the closed-form solution of Eq.~(\ref{eqn:obj_em}) is not given;
thus we optimize the objective by Newton's method in the M-step.
Recall that our goal is to divide the demonstration into several task phases; thus, 
we limit the state transitions to be unidirectional (or \textit{left-to-right}), i.e., $s_{1}=1 \leq s_{2} \leq \cdots \leq s_{T}=M$.

\subsection{Multi-objective Bayesian Optimization}
By involving $S$ determined by the segmentation, we subsequently conduct a multi-objective optimization of Eq.~(\ref{eqn:mult_obj_opt}) in real-world environments
using the previously proposed state-of-the-art Bayesian optimization method~\cite{ozaki2020multiobjective}. 
In addition, we exploit the identified stiffness $\theta^{\pi}=\{\stiffness^{\pi}_{j}\}$, through the previous IC-SLD step as a prior for the optimization by $\pi$-BO.
Thus, we define the prior $\pi(\theta)$ in Eq.~(\ref{eqn:pibo}) as follows:
\begin{align}
  \pi(\theta) &\coloneqq \prod_{j=1}^{M} \mathcal{N}(\stiffness_{j}; \stiffness^{\pi}_{j}, \sigma_{j}), \label{eqn:pi_def} \\
  \sigma_{j} &\coloneqq \min(\stiffness_{\mathrm{max}} - \stiffness^{\pi}_{j}, \stiffness^{\pi}_{j} - \stiffness_{\mathrm{min}}),
\end{align}
where $\stiffness_{\mathrm{max}}$, $\stiffness_{\mathrm{min}}$ are the maximum and minimum stiffness specified in the target system, respectively.

\subsection{Implementation Notes}
\begin{algorithm}[h]
\small
\DontPrintSemicolon
  \KwInput{
    A demonstrated trajectory $\tau$, \\
    Number of segmentations $M$, \\
    Number of optimization iterations $N$, \\
    Hyperparameters: $\beta$, $\kappa$
  }
  \KwOutput{Pareto solutions $\theta_{1:N}^{*}$}
  \tcp*[l]{(1) Segmentation of $\tau$ by IC-SLD}
  Optimize $\mathcal{J}_{\mathrm{EM}}$ defined by Eqs.~(\ref{eqn:obj_em}), (\ref{eqn:em_obj_jt3})
  to find ${\theta^{\pi}} = \{{K}^{\pi}_{j}\}$ and $S=\{s_{t}\}$. \\
  \tcp*[l]{(2) Multi-obj. optimization with $\pi$-BO}
  Define the prior $\pi(\theta)$ by Eq.~(\ref{eqn:pi_def}) \\
  Initialize dataset $\mathcal{D}_{0} \leftarrow \emptyset$ \\
  \For{$n \leftarrow 1$ \KwTo $N$}{
    Choose a candidate by $\alpha_{\pi}$ defined by Eqs.~(\ref{eqn:alpha_multi_obj}), (\ref{eqn:hypervolume}), (\ref{eqn:pibo}):
    $\theta_{n} = \argmax \alpha_{\pi}(\theta;\mathcal{D}_{n})$ \\
  Eval.{} the compliance objective: $y_{C} \leftarrow \mathcal{J}_{C}(\theta_{n}|S)$ \\
  Eval.{} the task objective in a real environment: $y_{T} \leftarrow \mathcal{J}_{T}(\theta_{n}|S)$ \\
  Update the dataset: $\mathcal{D}_{n} \leftarrow \mathcal{D}_{n-1} \cup \{(\theta_{n}, (y_{C}, y_{T}))\}$ \\
  }
  \Return $\theta_{1:N}^{*}$ in $\mathcal{D}_{N}$
\caption{T/C objectives optimization with IC-SLD and $\pi$-BO} \label{alg:isld_pibo}
\end{algorithm}

The pseudocode for the proposed algorithm is presented in Alg.~\ref{alg:isld_pibo}, and we present the following implementation details.
($\ell$1) We collect demonstrations at a sampling frequency of 20~Hz and perform IC-SLD segmentations at the same sampling frequency, i.e., $\Delta t =50$~ms.
Here, we assume that the stiffness matrices are diagonal.
($\ell$7) The task objective is then evaluated by actuating the robot
with the control parameters $\stiffness_{s_{1:T}}$, $\obs_{d,1:T}$,
where $\obs_{d,1:T}$ is the attractor trajectory to be tracked which is determined from $\stiffness_{s_{1:T}}$ and the demonstrated trajectory $\obs_{1:T}$, $\force_{1:T}$ as follows:
\begin{align}
  \obs_{d,t} = \obs_{t} + \stiffness^{-1}_{s_{t}} (2 \stiffness_{s_{t}}^{1/2}\vel_{t} + \inertia \ddot{\obs}_{t} - \force_{t}).
\end{align}
The above parameters are input to the feedback controller at 20~Hz, and the internal feedback operations are performed at a higher frequency (e.g., 1,000~Hz).

\section{Experiments} \label{sec:experiments}
We evaluated the effectiveness of the proposed method through simulation and real robot experiments.
In the following, we first specify the task settings and baselines in Secs.~\ref{sec:task_setting} and \ref{sec:baselines}, respectively. We then report the results of the simulation and real experiments in Secs.~\ref{sec:sim_eval} and \ref{sec:real_eval}, respectively.

\subsection{Task Settings} \label{sec:task_setting}
\begin{figure}
  \centering
  \includegraphics[width=0.45\textwidth]{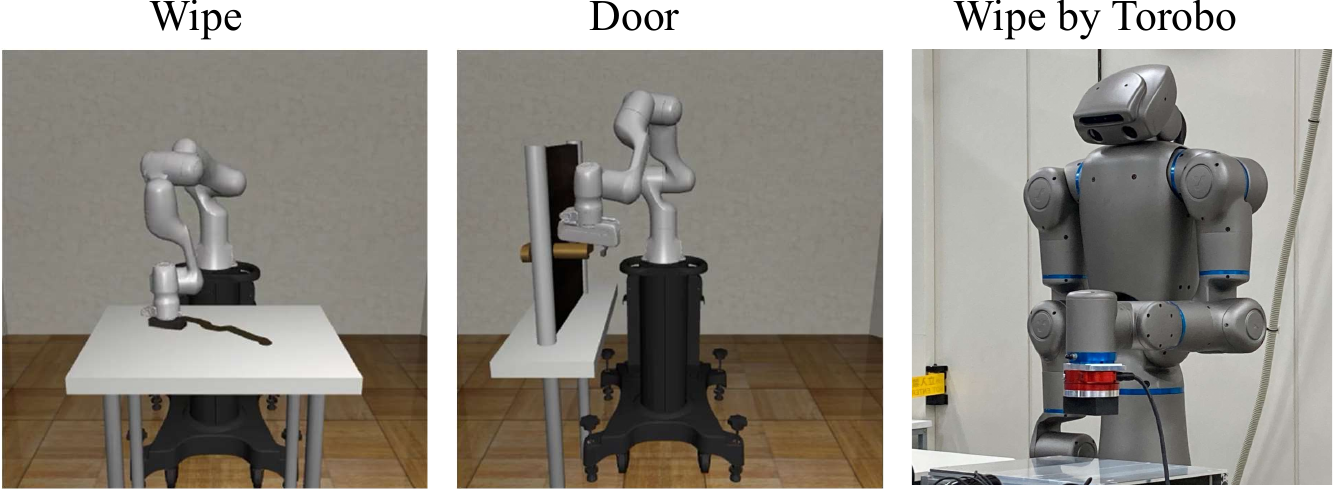}
  \caption{
    \texttt{Wipe} and \texttt{Door} simulation tasks in robosuite~\cite{robosuite2020},
    and a task on the Torobo.
    The rewards of the simulated tasks are binary variables indicating success or failure,
    where `1' indicates that the current state is a task completion state (dirt is cleaned or door is opened).
    The reward for the real task is a negative square error between the realized and demonstrated position trajectories.
  } \label{fig:robosuite}
\end{figure}

We employed two simulated tasks and a task using a real robot.
The simulated tasks included \texttt{Wipe} and \texttt{Door} from the \textit{robosuite} simulation framework~\cite{robosuite2020}.
We also performed the wipe task in real-world environments using the TokyoRobotics Torobo%
\footnote{\url{https://robotics.tokyo/products/torobo/}}.
The task objectives of these tasks were the sums of the task-specific reward function $R(\obs_{t})$.
Visualization of the experimental tasks and the reward definitions are summarized in Fig.~\ref{fig:robosuite}.
Here, demonstrations were collected by using a 3Dconnexion SpaceNavigator%
\footnote{
  \url{https://3dconnexion.com/jp/product/spacemouse-compact/}
}
in the simulator or teaching the Torobo directly.
For the \texttt{Door} task, the gripper actions involve the simple playback of the demonstrations.
Here, we set the IC-SLD hyperparameter to $\kappa=10^{-5}$ for the simulated tasks and
$\kappa=10^{-7}$ for the real task, based on the maximum acceptable stiffness of the systems.

\subsection{Baseline Methods} \label{sec:baselines}
The following two methods were selected as baselines for segmentation.

\subsubsection{Gaussian Mixture Model}
We performed GMM model fitting using $T$ samples from demonstrations. The GMM model is specified as follows:
\begin{align}
  p(\xi_{t}) = \sum^{M}_{j=1} \eta_{j} \mathcal{N}(\xi_{t}|\mu_{j}, \Sigma_{j}),
\end{align}
where $\xi_{t} \coloneqq (\obs_{t}, \vel_{t}, \ddot{\obs}_{t}, \force_{t})$, $\eta_{j}$ is the mixture coefficients, and $(\mu_{j}, \Sigma_{j})$ are parameters for Gaussian distributions.
After the fitting process, the segment identifiers were determined as
$s_{t} = \mathop{\rm argmax}_{j} \eta_{j} \mathcal{N}(\xi_{t};\mu_{j},\Sigma_{j})$.

\subsubsection{Switching Linear Dynamics unaware of Impedance Control}
We implemented an SLD-based baseline method unaware of impedance control,
i.e., this method does not exploit the priors of impedance control.
Here, the impedance control unaware linear dynamics is specified as:
$\state_{t} = \vel_{t}$, $\action_{t} = (\Delta \obs_{t}, \force_{t})$, and
\begin{align}
  A_{j} &= \mathrm{diag}(a_{1}, a_{2}, \cdots, a_{6}) \in \mathbb{R}^{6\times 6}, \\
  B_{j} &= \left(\mathrm{diag}(b_{1}, b_{2}, \cdots, b_{6}), \mathrm{diag}(b'_{1}, b'_{2}, \cdots, b'_{6})\right) \in \mathbb{R}^{6\times 12}.
\end{align}
With this formulation, we estimated the dynamics parameters and segment identifiers
by optimizing Eq.~(\ref{eqn:obj_em}) using the EM algorithm.

The above two methods are hereafter referred to as GMM and SLD.
Unless otherwise specified, $\pi$-BO was not applied to the baseline methods during the Bayesian optimization process.

\subsection{Simulation Evaluation} \label{sec:sim_eval}
The effectiveness of the proposed method was investigated through simulations.
We also evaluated the sensitivity to parameter settings.
For each setting, we conducted ten simulated trials with different random seeds,
and then we compared the results with the statistics (i.e., the means and standard deviations).

\subsubsection{Comparison with the baselines}
\begin{figure}[t]
  \centering
  \includegraphics[width=0.45\textwidth]{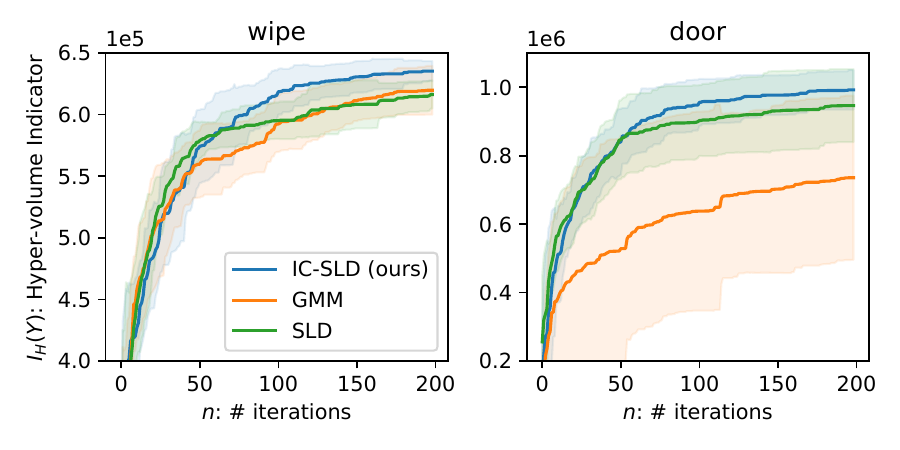}
  \vspace*{-3mm}
  \caption{
    Learning curves showing the growth of the hypervolume indicator $I_{H}(Y_{1:n}^{*})$.
    Solid lines and shaded areas show the means and standard deviations
    of the multiple trials, respectively.
  }
  \label{fig:learning_curve}
\end{figure}
Figure \ref{fig:learning_curve} shows the optimization progress obtained by the proposed method and baseline methods, where the hypervolume indicator $I_{H}(Y^{*}_{1:n})$ is used as the metric.
For \texttt{Wipe} and \texttt{Door}, $M$ was set to be $M=2$ and $M=3$, respectively.
In addition, the hyperparameter $\beta$ of $\pi$-BO was set to be $\beta=1$ for the proposed method.
As shown in Fig.~\ref{fig:learning_curve}, the proposed method optimized the metrics most efficiently,
achieving convergence within approximately 100 trials, corresponding to approximately one hour of learning%
\footnote{
A single episode takes within the 30s.
}%
.

\begin{figure}
  \centering
    \includegraphics[width=0.45\textwidth]{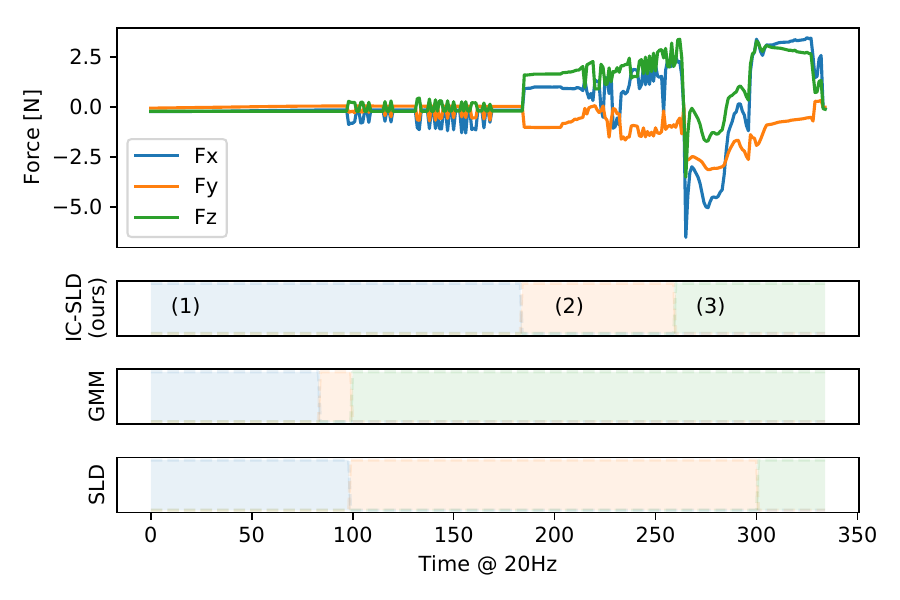}\\
    \includegraphics[width=0.35\textwidth]{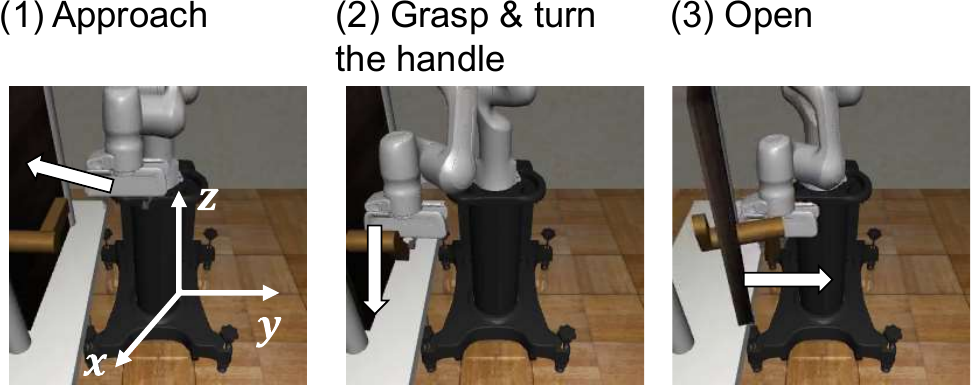}
  \caption{
    (Top) The demonstrated trajectory of the \texttt{Door} task, where only forces are shown for brevity,
    (middle) segmentation results obtained by IC-SLD, GMM, and SLD, where the different phases are color-coded, and
    (bottom) visualization of task phases found by proposed IC-SLD.
  }
  \label{fig:door_segs}
\end{figure}

Fig.~\ref{fig:door_segs} shows the segmentation results of the compared methods for the \text{Door} task.
As can be seen, the proposed IC-SLD method successfully found three interpretable phases of the task, i.e., (1) approach, (2) turn handle, and (3) open,
which indicates that this segmentation can improve the compliance objective by assigning a lower stiffness to phase (1).

\subsubsection{Ablation study}
This analysis was conducted to clarify which components of the proposed method (IC-SLD and $\pi$-BO)
contributed to the above improvement.
For this purpose, variants of the proposed method and baselines were prepared, i.e.,
the proposed method \textit{without} $\pi$-BO, and baselines \textit{with} $\pi$-BO.
For the baseline variants, the prior $\{\stiffness^{\pi}_{j}\}$ was computed by optimizing Eq.~(\ref{eqn:obj_em}) with fixed segmentation results obtained by GMM and SLD.
Table~\ref{tab:ablation} summarizes the ablation analysis,
demonstrating that involving both IC-SLD and $\pi$-BO contributed to performance improvement.
\begin{table*}
  \caption{
    Ablation Study. Hypervolume indicators ($\times 10^{3}$, Mean $\pm$ Std) at $n=100$.
    Bold and underlined values indicate the best results, and underlined values show the second-best results.
    } \label{tab:ablation}
  \centering
  \begin{tabular}[tb]{c|cccccc}
    \toprule
    Segmentation & \multicolumn{2}{c}{IC-SLD} & \multicolumn{2}{c}{GMM} & \multicolumn{2}{c}{SLD} \\
    $\pi$-BO ($\beta=1$) & $\checkmark$ & & $\checkmark$ & & $\checkmark$ & \\\midrule
    Wipe & \best{618.58}{18.49} & \third{593.98}{22.67} & \third{590.04}{22.74} & \third{592.40}{22.76} & \second{596.31}{38.38} & \third{595.12}{14.75} \\
    Door & \best{955.92}{68.54} & \second{923.05}{82.57} & \third{675.97}{319.80} & \third{638.04}{290.82} & \third{904.02}{131.36} & \third{902.80}{103.24} \\\bottomrule
  \end{tabular}
\end{table*}

\subsubsection{Parameter Sensitivity}
\begin{figure}
  \centering
  \includegraphics[width=0.45\textwidth]{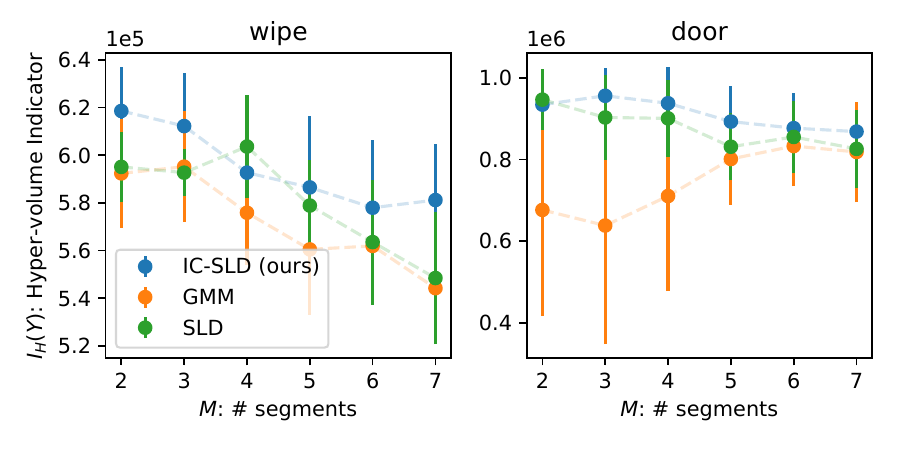} \\
  \includegraphics[width=0.45\textwidth]{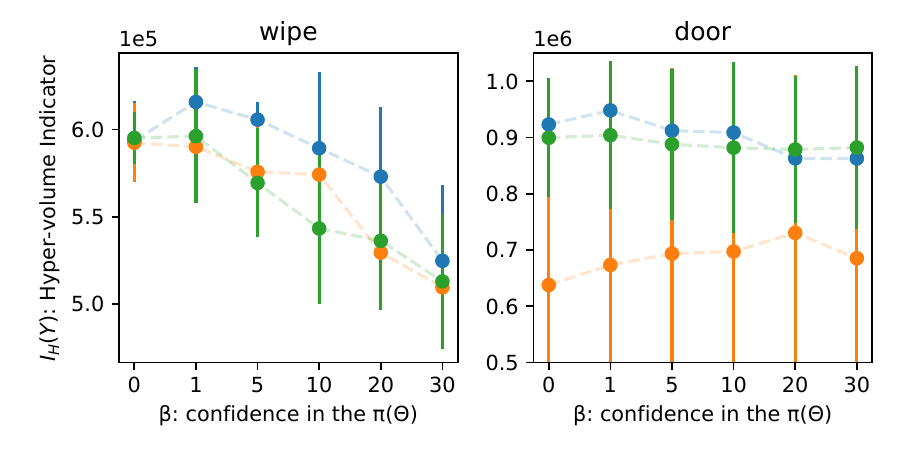}
  \vspace*{-3mm}
  \caption{Sensitivity analysis of hyperparameters $M$ and $\beta$.
  The error bars indicate the means and standard deviations at $n=100$.} \label{fig:sensitivity_analysis}
\end{figure}

Figure \ref{fig:sensitivity_analysis} summarizes the
sensitivity analysis for the hyperparameters $M$ and $\beta$.
As can be seen, $M=2$ and $M=3$ performed best on the two tasks for the proposed method,
and insufficient or too many divisions resulted in reduced performance.
Although increasing $M$ contributes to the expressiveness of the control,
optimization becomes increasingly difficult as the dimensionality of the input increases.
In addition, Fig.~\ref{fig:sensitivity_analysis} shows that, although optimization performance can be improved using the priors by setting $\beta > 0$,
overconfidence in the priors (i.e., higher $\beta$) led the optimizer to a local optimum.

\subsubsection{Tradeoff Analysis}
\begin{figure*}
  \centering
  \includegraphics[width=0.45\textwidth]{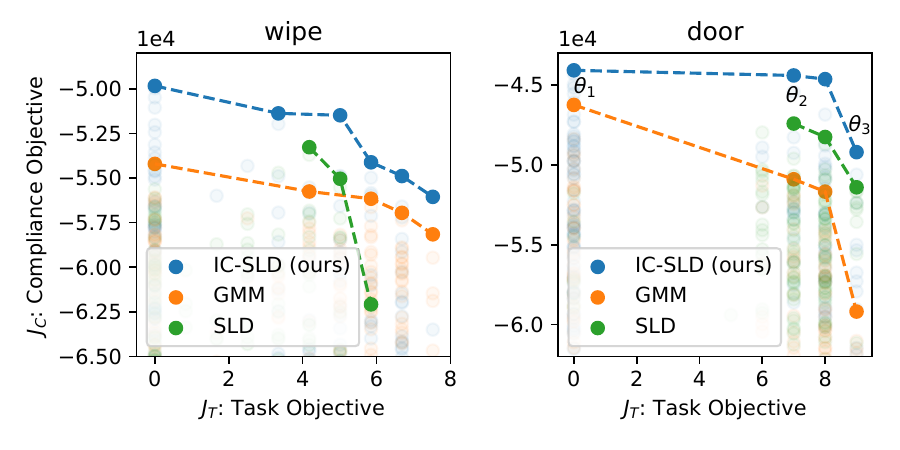}
  \includegraphics[width=0.3\textwidth]{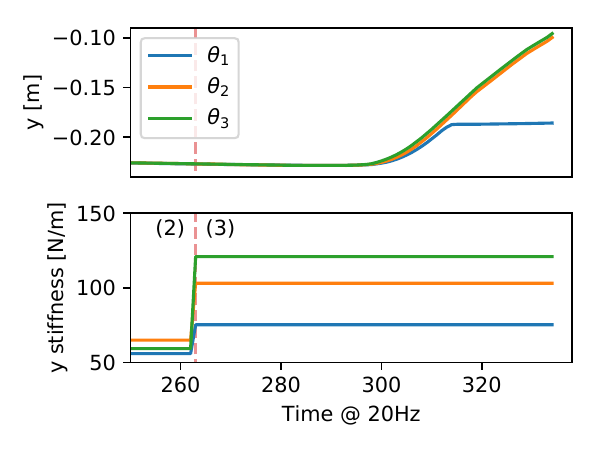}
  \vspace*{-3mm}
  \caption{
    (Two left) Pareto solutions (plotted non-transparently) for two simulated tasks, and (Right)
    behaviors of the different Pareto solutions on the \texttt{Door} task.
    For brevity, we showed only the behaviors related to the $y$-axis
    and focused on the phases of (2) and (3) illustrated in Fig.~\ref{fig:door_segs}.
  } \label{fig:door_tradeoff}
\end{figure*}

Figure~\ref{fig:door_tradeoff} shows the Pareto solutions obtained by the optimization presented in Fig.~\ref{fig:door_segs},
indicating that the proposed method obtained better Pareto solutions.
We also show the behaviors realized by different Pareto solutions $\theta_{1,2,3}$ on the \texttt{Door} task.
While the operation with $\theta_{1}$ failed to open the door due to its low stiffness,
the operations with $\theta_{2,3}$ could open the door with strong force produced by higher stiffness. 
In addition, although $\theta_{3}$ opened the door somewhat faster than $\theta_{2}$, the difference was negligible. 

\subsection{Evaluation in a Real-world Environment} \label{sec:real_eval}
\begin{figure}
  \centering
  \includegraphics[width=0.475\textwidth]{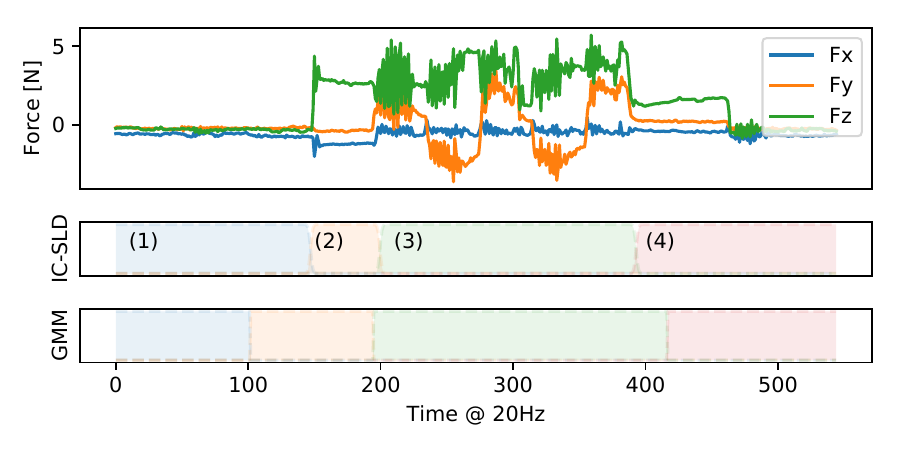}
  \includegraphics[width=0.475\textwidth]{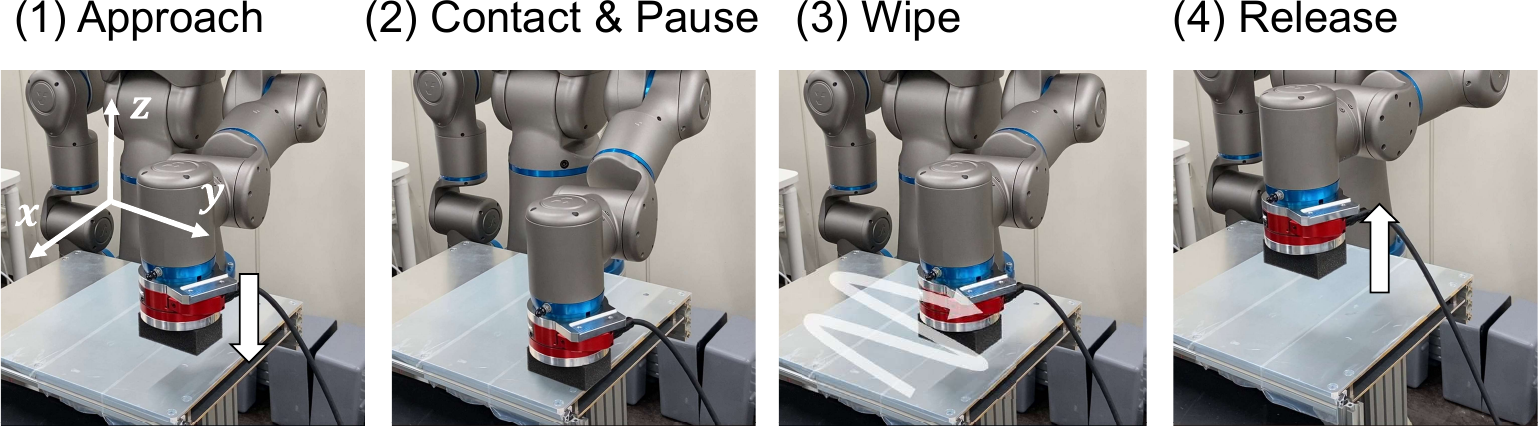}
  \caption{
    (Top) The demonstrated trajectory of the real Wipe task, where only forces are shown for brevity,
    (middle) segmentation results by IC-SLD and GMM, where different phases are color-coded, and
    (bottom) visualization of task phases found by the proposed IC-SLD.
    }
    \label{fig:torobo_seg_vis}
\end{figure}

\begin{figure*}
  \centering
  \includegraphics[width=0.475\textwidth]{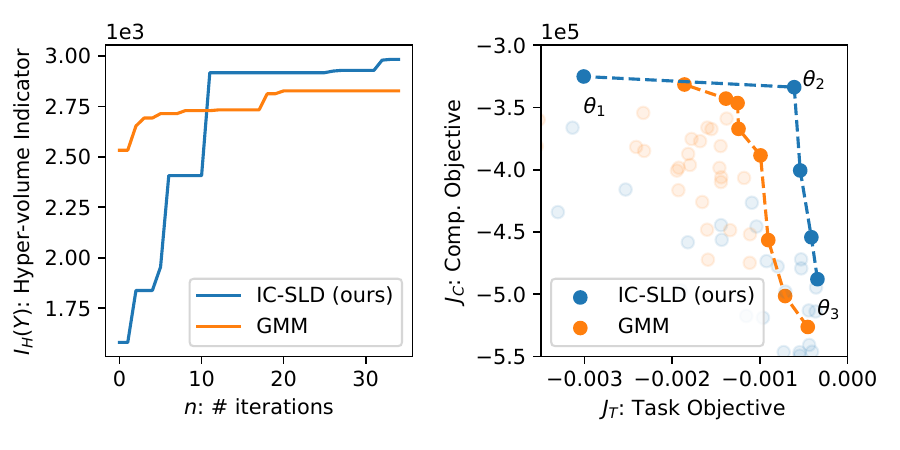}
  \includegraphics[width=0.475\textwidth]{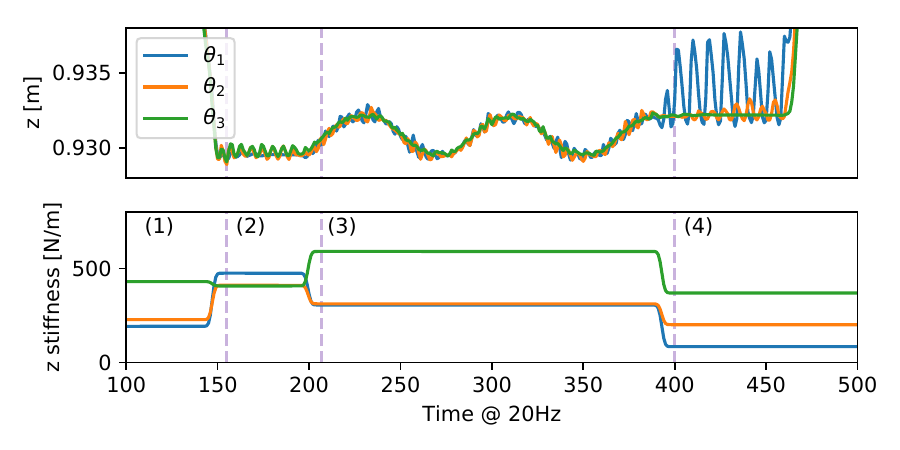}
  \vspace*{-3mm}
  \caption{Summary of the real Wipe task results; (left) learning curve, (middle) Pareto solutions, and (right) behaviors of different Pareto solutions.
  For brevity, we show only the behaviors related to the z-axis.}
  \label{fig:torbo_results}
\end{figure*}

An evaluation was conducted to verify the proposed method in a real-world environment.
In addition, a comparison was performed using GMM as a baseline method.
Figure \ref{fig:torobo_seg_vis} shows the segmentation results.
As can be seen, the proposed IC-SLD successfully identified interpretable task phases.
Figure \ref{fig:torbo_results} summarizes the experimental results,
highlighting that the proposed method found better Pareto solutions than the baseline method even in the real world.
The results of a tradeoff analysis (Fig.~\ref{fig:torbo_results}, right) demonstrated that high stiffness reduced the oscillation in the $z$-axis.
Based on this analysis, users can select which solution to utilize regarding acceptable oscillation and compliance.
%

\section{Conclusion} \label{sec:conclusion}
This paper has proposed a novel stiffness learning method to safely reproduce a human demonstration with impedance control.
Considering task and compliance objectives, the proposed method optimizes the stiffness parameters using multi-objective Bayesian optimization.
The proposed IC-SLD determines the search space for the Bayesian optimization, which effectively segments the demonstration into task phases suitable for switching stiffness impedance control.
In addition, the optimization is performed using the prior parameters obtained through the IC-SLD model identification.
The proposed method was evaluated experimentally using both simulated and real-world  robot tasks, and results demonstrate that the IC-SLD-based segmentation and prior utilization significantly improved optimization efficiency compared with previous baseline methods.

In this study, we assumed the stiffness matrix to be diagonal;
however, the effectiveness of nondiagonal stiffness matrices has been demonstrated in recent studies~\cite{oikawa2021reinforcement,kozlovsky2022reinforcement}.
Thus, it would be interesting to extend the proposed method to the nondiagonal setting by scaling the Bayesian optimization in high-dimensional input space.
Another important research topic is automatically determining the optimal number of segments $M$.
For this purpose, the nonparametric Bayesian inference~\cite{fox2011bayesian,villecroze2022bayesian} could be a promising approach. In addition, joint optimization of $M$ and stiffness is an attractive alternative.

\bibliography{iros2023}
\bibliographystyle{ieeetr}

\end{document}